\documentclass[a4paper]{article}

\usepackage{INTERSPEECH2016}

\usepackage{times}
\usepackage{url}
\usepackage{latexsym}

\usepackage{amssymb,amsmath,bm}
\usepackage{textcomp}
\usepackage{amsthm}

\usepackage{array}
\usepackage{color}

\usepackage{algorithmicx}
\usepackage{algpseudocode}
\renewcommand{\algorithmicrequire}{\textbf{Input:}}
\renewcommand{\algorithmicensure}{\textbf{Output:}}

\usepackage{graphicx}
\graphicspath{{figs/}}

\usepackage{multirow}

\newcommand{\figref}{Fig.\ref}
\newcommand{\equref}{Eq.\ref}
\newcommand{\tabref}{Tab.\ref}
\newcommand{\secref}{\ref}

\newcommand{\wg}{w}
\newcommand{\cg}{c}
\newcommand{\wj}{ws}
\newcommand{\cj}{cs}
\newcommand{\ws}{wsh}
\newcommand{\cs}{csh}
\newcommand{\cpw}{cpw}
\newcommand{\cpwj}{cpw}
\newcommand{\tied}{tied}

\ninept

\def\oneofall{oneofall}

\title{Model Interpolation with Trans-dimensional Random Field Language Models for Speech Recognition}

\makeatletter
\def\name#1{\gdef\@name{#1\\}}
\makeatother \name{{\em Bin Wang$^1$, Zhijian Ou$^1$, Yong He$^2$, Akinori Kawamura$^2$}}

\address{$^1$Department of Electronic Engineering, Tsinghua University, Beijing 100084, China \\
  $^2$Toshiba Corporation \\
  {\small \tt contact email: ozj@tsinghua.edu.cn}
}

\begin{document}

\maketitle

{
    \let\thefootnote\relax\footnotetext[0]{
        This work is supported by NSFC grant 61473168, Tsinghua Initiative 20121088069 and Toshiba Corporation.
    }
}

%\setlength{\abovecaptionskip}{2pt}
%\setlength{\belowcaptionskip}{-10pt}

%auto-ignore
\ifx\oneofall\undefined
\documentclass[a4paper]{article}

\usepackage{INTERSPEECH2016}

\usepackage{times}
\usepackage{url}
\usepackage{latexsym}

\usepackage{amssymb,amsmath,bm}
\usepackage{textcomp}
\usepackage{amsthm}
\newtheorem{thm}{Theorem}
\newtheorem{prop}{Proposition}
\usepackage{array}
\usepackage{color}

%peeudocoad
\usepackage{algorithmicx}
\usepackage{algpseudocode}
\renewcommand{\algorithmicrequire}{\textbf{Input:}}
\renewcommand{\algorithmicensure}{\textbf{Output:}}

%grpahic
\usepackage{graphicx}
\graphicspath{{figs/}}

%table
\usepackage{multirow}

% correct bad hyphenation here
\hyphenation{op-tical net-works semi-conduc-tor}

\newcommand{\figref}{Fig.\ref}
\newcommand{\equref}{Eq.\ref}
\newcommand{\tabref}{Tab.\ref}
\newcommand{\secref}{\ref}

\newcommand{\wg}{w}
\newcommand{\cg}{c}
\newcommand{\wj}{ws}
\newcommand{\cj}{cs}
\newcommand{\ws}{wsh}
\newcommand{\cs}{csh}
\newcommand{\cpw}{cpw}
\newcommand{\cpwj}{cpw}
\newcommand{\tied}{tied}

\sloppy % better line breaks
\ninept

\begin{document}
\fi

\begin{abstract}
The dominant language models (LMs) such as n-gram and neural network (NN) models represent sentence probabilities in terms of conditionals. In contrast, a new trans-dimensional random field (TRF) LM has been recently introduced to show superior performances, where the whole sentence is modeled as a random field.
In this paper, we examine how the TRF models can be interpolated with the NN models, and obtain 12.1\% and 17.9\% relative error rate reductions over 6-gram LMs for English and Chinese speech recognition respectively through log-linear combination.
		
\end{abstract}
\noindent{\bf Index Terms}: Language modeling, random fields, stochastic approximation

\section{Introduction}

Language modeling (LM) involves determining the joint probability of words in a sentence. The conditional approach is dominant, representing the joint probability in terms of conditionals. Examples include n-gram LMs \cite{chen1999empirical} and neural network (NN) LMs \cite{schwenk2007continuous,mikolov2011}.
Recently a new trans-dimensional random field (TRF) LM \cite{Bin2015} is introduced, where the whole sentence is modeled as a random field.
As the random field approach avoids local normalization which is required in the conditional approach, it is computationally more efficient in computing sentence probabilities and has the potential advantage of being able to flexibly integrate a richer set of features.
An effective training algorithm using joint stochastic approximation (SA) and trans-dimensional mixture sampling is developed in \cite{Bin2015}.
It is found that the TRF models significantly outperformed the modified Kneser-Ney (KN) smoothed 4-gram LM with 9.1\% relative reduction in speech recognition word error rates (WERs) and performed slightly better than the recurrent neural network LMs but with 200x faster speed in re-scoring n-best lists of hypothesized sentences. To our knowledge, this result represents the first strong empirical evidence supporting the power of using the whole-sentence random field approach for LMs \cite{rosenfeld2001whole}.

In this paper, we examine how the TRF models can be interpolated with NN models to further improve the performance.
Speech recognition experiments are conducted for both English and Chinese.
Three kinds of interpolation schemes, namely linear interpolation (word-level or sentence-level) and log-linear interpolation, are evaluated for model combinations between KN n-gram LMs, NN LMs and our TRF LMs.
Through log-linearly combining TRF LMs with NN LMs, we obtain 12.1\% and 17.9\% relative error rate reductions over the KN 6-gram LMs for English and Chinese speech recognition respectively.

\ifx\oneofall\undefined
\bibliographystyle{IEEEtran}
\bibliography{RF}
\end{document}
\fi

%auto-ignore
\ifx\oneofall\undefined
\documentclass[a4paper]{article}

\usepackage{INTERSPEECH2016}

\usepackage{times}
\usepackage{url}
\usepackage{latexsym}

\usepackage{amssymb,amsmath,bm}
\usepackage{textcomp}
\usepackage{amsthm}
\newtheorem{thm}{Theorem}
\newtheorem{prop}{Proposition}
\usepackage{array}
\usepackage{color}

%peeudocoad
\usepackage{algorithmicx}
\usepackage{algpseudocode}
\renewcommand{\algorithmicrequire}{\textbf{Input:}}
\renewcommand{\algorithmicensure}{\textbf{Output:}}

%grpahic
\usepackage{graphicx}
\graphicspath{{figs/}}

%table
\usepackage{multirow}

% correct bad hyphenation here
\hyphenation{op-tical net-works semi-conduc-tor}

\newcommand{\figref}{Fig.\ref}
\newcommand{\equref}{Eq.\ref}
\newcommand{\tabref}{Tab.\ref}
\newcommand{\secref}{\ref}

\newcommand{\wg}{w}
\newcommand{\cg}{c}
\newcommand{\wj}{ws}
\newcommand{\cj}{cs}
\newcommand{\ws}{wsh}
\newcommand{\cs}{csh}
\newcommand{\cpw}{cpw}
\newcommand{\cpwj}{cpw}
\newcommand{\tied}{tied}

\sloppy % better line breaks
\ninept

\begin{document}
\fi

\section{Background of TRF Model Definition and Training}\label{sec:model}

    Throughout, we denote by $x^l=(x_1,\ldots, x_l)$ a sentence (i.e., word sequence) of length $l$ ranging from 1 to $m$.
    Each element of $x^l$ corresponds to a single word.
    $D$ denotes the whole training corpus and $D_l$ denotes the collection of length $l$ in the training corpus.
    $n_l$ denotes the size of $D_l$ and $n=\sum_{l=1}^{m} n_l$.

    As defined in \cite{Bin2015},
    a trans-dimensional random field model represents the joint probability of the pair $(l, x^l)$ as
    \begin{equation}\label{eq:model}
        p(l,x^l;\lambda) = \frac{n_l/n}{Z_l(\lambda)} e^{\lambda^T f(x^l)},
    \end{equation}
    where $n_l/n$ is the empirical probability of length $l$.
%    $p_l(x^l;\lambda)$ is the probability of $x^l$ given the length $l$, which is defined to be a random field model:
%    \begin{equation}\label{eq:sub-model}
%        p_l(x^l; \lambda) = \frac{1}{Z_l(\lambda)} e^{\lambda^T f(x^l)},
%    \end{equation}
%    In \equref{eq:sub-model},
    $f(x^l)=(f_1(x^l), \ldots f_d(x^l))^T$ is the feature vector,
    which is usually defined to be position-independent and length-independent, e.g. the $n$-grams.
    $d$ is the dimension of the feature vector $f(x)$.
    $\lambda$ is the corresponding parameter vector of $f(x^l)$.
    $Z_l(\lambda) = \sum_{x^l} e^{\lambda^T f(x^l)}$ is the normalization constant of length $l$.
    By making explicit the role of length in model definition, it is clear that the model in \eqref{eq:model} is a mixture of random fields on sentences of different lengths (namely on subspaces of different dimensions), and hence will be called a trans-dimensional random field (TRF).

In the joint SA training algorithm \cite{Bin2015}, another form of mixture distribution is defined as follows:
    \begin{equation} \label{eq:model-full}
    p(l, x^l; \lambda, \zeta) = \frac{n_l/n}{Z_1(\lambda) e^{\zeta_l}} e^{\lambda^T f(x^l)}
    \end{equation}
    where $\zeta=\{\zeta_1, \dots, \zeta_m\}$ with $\zeta_1=0$ and $\zeta_l$ is the hypothesized value of the log ratio of $Z_l(\lambda)$ with respect to $Z_1(\lambda)$, namely $\log \frac{Z_l(\lambda)}{Z_1(\lambda)}$.
    $Z_1(\lambda)$ is chosen as the reference value and can be calculated exactly.
An important observation is that if and only if $\zeta$ were equal to the true log ratios, then the marginal probability of length $l$ under distribution \eqref{eq:model-full} equals to $n_l/n$. This property is then used to construct the joint SA algorithm, which jointly estimates the model parameters and normalization constants. More recent development on TRF models and related algorithms, see \cite{Bin2016}.

\ifx\oneofall\undefined
\bibliographystyle{IEEEtran}
\bibliography{RF}
\end{document}
\fi

%auto-ignore
\ifx\oneofall\undefined
\documentclass[a4paper]{article}

\usepackage{INTERSPEECH2016}

\usepackage{times}
\usepackage{url}
\usepackage{latexsym}

\usepackage{amssymb,amsmath,bm}
\usepackage{textcomp}
\usepackage{amsthm}
\newtheorem{thm}{Theorem}
\newtheorem{prop}{Proposition}
\usepackage{array}
\usepackage{color}

%peeudocoad
\usepackage{algorithmicx}
\usepackage{algpseudocode}
\renewcommand{\algorithmicrequire}{\textbf{Input:}}
\renewcommand{\algorithmicensure}{\textbf{Output:}}

%grpahic
\usepackage{graphicx}
\graphicspath{{figs/}}

%table
\usepackage{multirow}

% correct bad hyphenation here
\hyphenation{op-tical net-works semi-conduc-tor}

\newcommand{\figref}{Fig.\ref}
\newcommand{\equref}{Eq.\ref}
\newcommand{\tabref}{Tab.\ref}
\newcommand{\secref}{\ref}

\newcommand{\wg}{w}
\newcommand{\cg}{c}
\newcommand{\wj}{ws}
\newcommand{\cj}{cs}
\newcommand{\ws}{wsh}
\newcommand{\cs}{csh}
\newcommand{\cpw}{cpw}
\newcommand{\cpwj}{cpw}
\newcommand{\tied}{tied}

\sloppy % better line breaks
\ninept

\begin{document}
\fi

\section{Experiments}

    \begin{table}
    \centering
    \begin{tabular}{l|l}
        \hline
        W  & $s(x^l) = \prod_{i=1}^{l} \left[ \alpha p_1(x_i|h_i) + (1-\alpha)p_2(x_i|h_i) \right]$ \\
        \hline
        S  & $s(x^l) = \alpha p_1(x^l) + (1-\alpha) p_2(x^l)$ \\
        \hline
        Log & $s(x^l) = \exp( \alpha \log{p_1(x^l)} + (1-\alpha) \log{p_2(x^l)} )$ \\
        \hline
    \end{tabular}
   % \vspace{-8pt}
    \caption{
    Model interpolation schemes.
    $s(x^l)$ is the combined score of sentence $x^l$.
    $p_1(x_i|h_i)$ and $p_2(x_i|h_i)$ are the conditional probabilities of $x_i$ given history $h_i$ estimated by two LMs;
    $p_1(x^l)$ and $p_2(x^l)$ are the joint probabilities of the whole sentence $x^l$ estimated by two LMs.
    $0 < \alpha < 1$ is the interpolation weight which is tuned on the development set.
    } \label{tab:inter}
   % \vspace{-10pt}
    \end{table}

We examine how the TRF models can be interpolated with NN models to further improve the performance. Linear combination can be done at either word-level (W) or sentence-level (S). In contrast, log-linear combination has no such differentiation. The three schemes are detailed in \tabref{tab:inter}. As TRF models only output sentence probabilities, the "W" scheme is not applicable in combining TRF models with other models.

\subsection{English speech recognition on WSJ0 dataset}\label{subsec:wsj}

\begin{table}
\centering
\small
%\vspace{-10pt}
\begin{tabular}{l|c|c|c}
    \hline
    Model   & WER   & PPL   & \#feat (M) \\
    \hline \hline
    %KN4 (cutoff=0000)            & 8.71  & 295.41    & 1.6  \\
    KN5 (cutoff=00000)            & 8.61  & 293.94    & 2.3  \\
    KN6 (cutoff=000000)            & 8.61  & 293.67    & 3.0  \\
    RNN \cite{mikolov2011}             & 7.96  & 256.15    & 5.1 \\
    \hline \hline
    \multicolumn{4}{l}{ TRFs (200 classes) } \\
    \hline
    %\wg+\cg+\wj+\cj           & 8.16 & $\sim$260 & 5.2 \\
    \wg+\cg+\wj+\cj+\cpw      & 7.92 & $\sim$257 & 6.4 \\
    \wg+\cg+\wj+\cj+\ws+\cs   & 7.94 & $\sim$259 & 5.9 \\
    \wg+\cg+\wj+\cj+\ws+\cs+\tied* & \textbf{7.85} & $\sim$261 & 7.7 \\
    \hline \hline
    \multicolumn{4}{l}{The above results are reproduced from \cite{Bin2016}.}\\
    \hline
    \multicolumn{4}{l}{ Model Combination } \\
    \hline
    RNN + KN5 (W)               & 7.85\ & & \\
    RNN + KN5 (S)               & 7.96 & & \\
    RNN + KN5 (Log)             & 7.96 & & \\
    TRF*+ RNN (S)               & \textbf{7.51} & & \\
    TRF*+ RNN (Log)             & \textbf{7.57} & & \\
    \hline
\end{tabular}
%    \vspace{-10pt}
    \caption{
        The WERs and PPLs on the WSJ'92 test data.
        ``\#feat'' denotes the feature size (million).
        ``(W),(S),(Log)'' denote different model combination schemes as defined in \tabref{tab:inter}.
        ``TRF*'' denotes the TRF with features ``\wg+\cg+\wj+\cj+\ws+\cs+\tied''.
        "\tied" represents the tied long-skip-bigram features introduced in \cite{Bin2016}, in which the skip-bigrams with skipping distances from 6 to 9 share the same parameter.        
        }
\label{tab:exp2}
%\vspace{-18pt}
\end{table}

    In this section, speech recognition and 1000-best list rescoring experiments are conducted as configured in \cite{Bin2015, Bin2016}.
    The maximum length of TRFs is $m=82$, which is equal to the maximum length of the training sentences.
    The other configurations are: $K=300$, $\beta_\lambda=0.8$, $\beta_\zeta=0.6$, $t_c=3000$, $t_0 = 2000$,
    $t_{max}=20,000$.
    L2 regularization with constant $4 \times 10^{-5}$ is used to avoid over-fitting.
    6 CPU cores are used to parallelize the algorithm.
    The word error rates (WERs) and perplexities (PPLs) on WSJ'92 test set are shown in \tabref{tab:exp2}.
	
    Combining TRF and KN5 provides no WER reduction.
    Different schemes give close WERs for combining TRF  and RNN.
	The log-linear interpolation performs more stable when considering both English and Chinese experiments (as shown later).
    For English, the obtained WER 7.57\% indicates 12.1\% and 3.6\% relative reductions, when compared to the result of using KN6 (8.61\%) and the best result of combining RNN and KN5 (7.85\%) respectively.

\subsection{Chinese speech recognition on Toshiba dataset}

    In this section we report the results from using TRF LMs in a large vocabulary Mandarin speech recognition experiment.
    Different LMs are evaluated by rescoring 30000-best list from a Toshiba's internal test set (2975 utterances).
    The oracle character error rate (CER) of the 30000-best lists is 1.61\%, which are generated with a DNN-based acoustic model.
    The LM corpus is from Toshiba, which contains about 20M words.
    We randomly select 1\% from the corpus as the development set and others as the training set.
    The vocabulary contains 82K words, with one special token $<$UNK$>$.
    The NN LM used here is the feedforward neural network (FNN) LM \cite{schwenk2007continuous, schwenk2013cslm} trained by CSLM toolkit\footnote{
    \url{http://www-lium.univ-lemans.fr/cslm/}
    }.
    The number of hidden units is 512 and the projection layer units is 3$\times$128.
    The TRF models are trained using the feature set ``\wg+\cg+\wj+\cj+\cpw'' with different numbers of classes (200,400,600).
    The configurations are: $m=100$, $\beta_\lambda=0.8$, $\beta_\zeta=0.6$, $t_c=1000$, $t_0=t_{max}=20000$.
    The sample number $K$ is increased from 300 until no improvements on the development set are observed, and finally set to be 8000.
%    The initial training configuration is: $K=300$, $\beta_\lambda=0.8$, $\beta_\zeta=0.6$, $t_c=1000$, $t_0=t_max=20000$. And we gradually increase the batch number $K$ and at last the batch number reach 8000.
    20 CPU cores are used to parallelize the algorithm.
    The CERs and PPLs on the test set are shown in \tabref{tab:exp3}.

\begin{table}
\centering
\small
\begin{tabular}{l|c|c|c}
    \hline
    Model   & CER   & PPL   & \#feat (M) \\
    \hline \hline
    %KN4 (cutoff=0000)     & 4.91  & 229.72    & 33  \\
    KN5 (cutoff=00000)    & 4.88  & 227.92    & 49  \\
    KN6 (cutoff=000000)   & 4.87  & 227.76    & 64  \\
    FNN \cite{schwenk2013cslm}         & 4.30  & 206.87    & 53 \\
    \hline \hline
    \multicolumn{4}{l}{TRFs (``\wg+\cg+\wj+\cj+\cpw'')}  \\
    \hline
    200 classes & 4.52 & $\sim$207 & 47 \\
    400 classes* & 4.32 & $\sim$191 & 58 \\
    600 classes & 4.35 & $\sim$186 & 61 \\
    \hline \hline
    \multicolumn{4}{l}{Model Combination} \\
    \hline
    FNN + KN6 (W) & 4.06 & & \\
    FNN + KN6 (S) & 4.16 & & \\
    FNN + KN6 (Log) & 4.11 & & \\
    TRF* + FNN (S) & 4.09 & & \\
    TRF* + FNN (Log) & \textbf{4.0} & & \\
    \hline
\end{tabular}
%\vspace{-9pt}
\caption{
The CERs and PPLs on the test set in Chinese speech recognition.
%``\#feat'' denotes the feature number (million).
``TRF*'' denotes the TRF trained with 400 classes.
}
\label{tab:exp3}
%\vspace{-17pt}
\end{table}

   The results again demonstrate that the TRF LMs significantly outperform the n-gram LMs and are able to match the NN LM.
   Log-linear combination of TRF and FNN further reduces the CER.
   The obtained CER 4.0\% indicates 17.9\% and 1.5\% relative reductions, when compared to the result of using KN6 (4.87\%) and the best result of combining FNN and KN6 (4.06\%) respectively.

\ifx\oneofall\undefined
\bibliographystyle{IEEEtran}
\bibliography{RF}
\end{document}
\fi

%auto-ignore
\ifx\oneofall\undefined
\documentclass[a4paper]{article}

\usepackage{INTERSPEECH2016}

\usepackage{times}
\usepackage{url}
\usepackage{latexsym}

\usepackage{amssymb,amsmath,bm}
\usepackage{textcomp}
\usepackage{amsthm}
\newtheorem{thm}{Theorem}
\newtheorem{prop}{Proposition}
\usepackage{array}
\usepackage{color}

%peeudocoad
\usepackage{algorithmicx}
\usepackage{algpseudocode}
\renewcommand{\algorithmicrequire}{\textbf{Input:}}
\renewcommand{\algorithmicensure}{\textbf{Output:}}

%grpahic
\usepackage{graphicx}
\graphicspath{{figs/}}

%table
\usepackage{multirow}

% correct bad hyphenation here
\hyphenation{op-tical net-works semi-conduc-tor}

\newcommand{\figref}{Fig.\ref}
\newcommand{\equref}{Eq.\ref}
\newcommand{\tabref}{Tab.\ref}
\newcommand{\secref}{\ref}

\newcommand{\wg}{w}
\newcommand{\cg}{c}
\newcommand{\wj}{ws}
\newcommand{\cj}{cs}
\newcommand{\ws}{wsh}
\newcommand{\cs}{csh}
\newcommand{\cpw}{cpw}
\newcommand{\cpwj}{cpw}
\newcommand{\tied}{tied}

\sloppy % better line breaks
\ninept

\begin{document}
\fi

\section{Related Work and Discussion}

Currently, most attention of LM research is attracted by using neural networks, which has been shown to surpass the classic n-gram LMs.
Two basic classes of NN LMs are based on FNN \cite{schwenk2007continuous} and RNN \cite{mikolov2011}.
Recent extensions involve the use of sum-product networks \cite{cheng2014language}, deep recurrent neural networks \cite{zaremba2014recurrent} and feedforward sequential memory networks \cite{zhang2015feedforward}; only perplexity results are reported in these studies.
Crucially, no matter what form the networks take, various NN LMs follow the conditional approach and thus suffer from the expensive softmax computations due to the requirement of local normalization.
Lots of studies aim to alleviate this deficiency. Initial efforts include
using hierarchical output layer structure with word clustering \cite{mikolov2011}, converting NNs to n-gram LMs \cite{arisoy2014converting}.
Recently a number of studies \cite{mnih2012fast, vaswani2013decoding, Unnormalized2015, chen2015recurrent} make use of noise contrastive estimation (NCE) \cite{NCE2012} to build unnormalized variants of NN LMs through trickily avoiding local normalization in training and heuristically fixing the normalizing term in testing.

In contrast, TRF LMs eliminate local normalization from the root and thus are much more efficient in testing with theoretical guarantee. Empirically it is found in \cite{Bin2015} that 
the average time costs for re-ranking of the 1000-best list for a sentence are 0.16 sec vs. 40 sec, based on TRF and RNN respectively (no GPU used).
Equally importantly, evaluations in this paper and also in \cite{Bin2015, Bin2016} have shown that TRF LMs are able to perform as good as NN LMs (either RNN or FNN) on a variety of tasks.
Encouragingly, the random field approach may open a new door to language modeling in addition to the dominant conditional approach, as once envisioned in \cite{rosenfeld2001whole}. Integrating richer features and introducing hidden variables are worthwhile future works.

%\section{Acknowledgements}
%\label{sec:ack}
%
%    This work is partly supported by Toshiba Corporation and Natural Science Foundation of China under grant No. 61075020.

\ifx\oneofall\undefined
\bibliographystyle{IEEEtran}
\bibliography{RF}
\end{document}
\fi

\newpage
\eightpt
\bibliographystyle{IEEEtran}
\bibliography{RF}

\begin{thebibliography}{10}
\providecommand{\url}[1]{#1}
\csname url@samestyle\endcsname
\providecommand{\newblock}{\relax}
\providecommand{\bibinfo}[2]{#2}
\providecommand{\BIBentrySTDinterwordspacing}{\spaceskip=0pt\relax}
\providecommand{\BIBentryALTinterwordstretchfactor}{4}
\providecommand{\BIBentryALTinterwordspacing}{\spaceskip=\fontdimen2\font plus
\BIBentryALTinterwordstretchfactor\fontdimen3\font minus
  \fontdimen4\font\relax}
\providecommand{\BIBforeignlanguage}[2]{{%
\expandafter\ifx\csname l@#1\endcsname\relax
\typeout{** WARNING: IEEEtran.bst: No hyphenation pattern has been}%
\typeout{** loaded for the language `#1'. Using the pattern for}%
\typeout{** the default language instead.}%
\else
\language=\csname l@#1\endcsname
\fi
#2}}
\providecommand{\BIBdecl}{\relax}
\BIBdecl

\bibitem{chen1999empirical}
S.~F. Chen and J.~Goodman, ``An empirical study of smoothing techniques for
  language modeling,'' \emph{Computer Speech \& Language}, vol.~13, pp.
  359--394, 1999.

\bibitem{schwenk2007continuous}
H.~Schwenk, ``Continuous space language models,'' \emph{Computer Speech \&
  Language}, vol.~21, pp. 492--518, 2007.

\bibitem{mikolov2011}
T.~Mikolov, S.~Kombrink, L.~Burget, J.~H. Cernocky, and S.~Khudanpur,
  ``Extensions of recurrent neural network language model,'' in \emph{Proc.
  ICASSP}, 2011.

\bibitem{Bin2015}
B.~Wang, Z.~Ou, and Z.~Tan, ``Trans-dimensional random fields for language
  modeling,'' in \emph{Proceedings of the Association for Computational
  Linguistics (ACL)}, 2015.

\bibitem{rosenfeld2001whole}
R.~Rosenfeld, S.~F. Chen, and X.~Zhu, ``Whole-sentence exponential language
  models: a vehicle for linguistic-statistical integration,'' \emph{Computer
  Speech \& Language}, vol.~15, pp. 55--73, 2001.

\bibitem{Bin2016}
B.~Wang, Z.~Ou, and Z.~Tan, ``Learning trans-dimensional random fields with
  applications to language modeling,'' \emph{Draft}, 2016.

\bibitem{schwenk2013cslm}
H.~Schwenk, ``Cslm - a modular open-source continuous space language modeling
  toolkit.'' in \emph{Proc. INTERSPEECH}, 2013, pp. 1198--1202.

\bibitem{cheng2014language}
W.-C. Cheng, S.~Kok, H.~V. Pham, H.~L. Chieu, and K.~M.~A. Chai, ``Language
  modeling with sum-product networks.'' in \emph{INTERSPEECH}, 2014.

\bibitem{zaremba2014recurrent}
W.~Zaremba, I.~Sutskever, and O.~Vinyals, ``Recurrent neural network
  regularization,'' \emph{arXiv preprint arXiv:1409.2329}, 2014.

\bibitem{zhang2015feedforward}
S.~Zhang, C.~Liu, H.~Jiang, S.~Wei, L.~Dai, and Y.~Hu, ``Feedforward sequential
  memory networks: A new structure to learn long-term dependency,'' \emph{arXiv
  preprint arXiv:1512.08301}, 2015.

\bibitem{arisoy2014converting}
E.~Arisoy, S.~F. Chen, B.~Ramabhadran, and A.~Sethy, ``Converting neural
  network language models into back-off language models for efficient decoding
  in automatic speech recognition,'' \emph{IEEE/ACM Transactions on Audio,
  Speech, and Language Processing}, vol.~22, no.~1, pp. 184--192, 2014.

\bibitem{mnih2012fast}
A.~Mnih and Y.~W. Teh, ``A fast and simple algorithm for training neural
  probabilistic language models,'' \emph{arXiv preprint arXiv:1206.6426}, 2012.

\bibitem{vaswani2013decoding}
A.~Vaswani, Y.~Zhao, V.~Fossum, and D.~Chiang, ``Decoding with large-scale
  neural language models improves translation.'' in \emph{EMNLP}, 2013, pp.
  1387--1392.

\bibitem{Unnormalized2015}
A.~Sethy, S.~Chen, E.~Arisoy, and B.~Ramabhadran, ``Unnormalized exponential
  and neural network language models,'' in \emph{Proc. ICASSP}, 2015.

\bibitem{chen2015recurrent}
X.~Chen, X.~Liu, M.~Gales, and P.~Woodland, ``Recurrent neural network language
  model training with noise contrastive estimation for speech recognition,'' in
  \emph{Proc. ICASSP}, 2015, pp. 5411--5415.

\bibitem{NCE2012}
M.~U. Gutmann and A.~Hyv\"arinen, ``Noise-contrastive estimation of
  unnormalized statistical models, with application to natural image
  statistics,'' \emph{The journal of Machine Learning Research}, vol.~13,
  no.~1, pp. 307--361, 2012.

\end{thebibliography}
\end{document}